\documentclass[11pt]{article}

\usepackage[final]{acl}

\usepackage{times}
\usepackage{latexsym}
\usepackage[T1]{fontenc}
\usepackage[utf8]{inputenc}
\usepackage{microtype}
\usepackage{inconsolata}

\usepackage{amsmath}
\usepackage{hyperref}
\usepackage{url}
\usepackage{booktabs}
\usepackage{amsfonts}
\usepackage{nicefrac}
\usepackage{xcolor}
\usepackage{graphicx}
\usepackage{enumitem}
\usepackage{multirow}
\usepackage{subcaption}
\usepackage{makecell}
\usepackage{array}
\usepackage{pifont}
\usepackage{amssymb}

\newcommand{\cmark}{\ding{51}}
\newcommand{\xmark}{\ding{55}}
\newcommand{\pmark}{\raisebox{0.15ex}{\scalebox{0.85}{$\blacktriangle$}}}

\title{PTCG-Bench: Can LLM Agents Master Pokémon Trading Card Game?}

\author{
 \textbf{Dongdong Hua\textsuperscript{1}},
 \textbf{Yifei Sun\textsuperscript{1}},
 \textbf{Renhong Huang\textsuperscript{1}},
 \textbf{Feng Gao\textsuperscript{2}},
\\
 \textbf{Chunping Wang\textsuperscript{2}},
 \textbf{Yang Yang\textsuperscript{1}\thanks{~~Corresponding author.}}
\\
 \textsuperscript{1} Zhejiang University,
 \textsuperscript{2} FinVolution Group
\\
 \texttt{\{ddhua,yifeisun,renh2,yangya\}@zju.edu.cn}
\\
 \texttt{\{gaofeng02,wangchunping02\}@xinye.com}
}

\begin{document}

\maketitle

\begin{abstract}
Given a strategically complex board game, human players can quickly learn to devise strategies after playing a few rounds. Autonomous agents require similar capabilities in realistic interactive environments, yet existing agent benchmarks often fail to fully capture such strategic and evolving decision-making scenarios. We present \textbf{PTCG-Bench}, a benchmark built on the Pok\'{e}mon Trading Card Game (PTCG) that evaluates LLM agents at two complementary levels: (1) their decision-making performance within a single complex environment, and (2) their ability to self-evolve through accumulated experience. We further include a modular harness ablation to better interpret agent performance without conflating it with model capability. Our experiments show that, although LLM agents can achieve non-trivial gameplay performance, sustained and stable self-evolution remains challenging, and performance is sensitive to harness design. We hope that PTCG-Bench will facilitate future research on harness-aware and self-evolving agents in realistic interactive environments.\footnote{Code is available at \href{https://github.com/zjunet/PTCG-Bench}{https://github.com/zjunet/PTCG-Bench}.}
\end{abstract}
\section{Introduction}

Games have long served as important testbeds for autonomous agents, driving progress from Atari~\citep{mnih2015human}, Go~\citep{silver2017mastering}, and StarCraft II~\citep{vinyals2019grandmaster} to recent LLM agents in Minecraft~\citep{wang2023voyager} and Diplomacy~\citep{meta2022human}. More recently, Pok\'{e}mon has attracted growing attention as a challenging domain for both competitive battling~\citep{karten2025pok} and open-ended RPG gameplay~\citep{anthropic2025visible}. Meanwhile, game-based benchmarks have expanded from classic environments such as ALE~\citep{bellemare2013arcade}, TextWorld~\citep{cote2018textworld}, ALFWorld~\citep{shridhar2020alfworld}, and MineDojo~\citep{fan2022minedojo} to LLM-agent benchmarks such as lmgame-Bench~\citep{hu2025lmgame}, Orak~\citep{park2025orak}, and PokeAgent Challenge~\citep{karten2026pokeagent}. However, as summarized in Table~\ref{tab:game_bench_comparison}, existing benchmarks often emphasize isolated aspects of agent capability, provide limited support for evaluating sustained self-evolution within the same strategic environment, or leave harness-level effects insufficiently separated from model capability. This leaves open the need for a unified benchmark that can jointly assess strategic gameplay, stable self-evolution, and harness-aware agent evaluation.

\begin{figure*}[t]
    \centering
    \includegraphics[width=\linewidth]{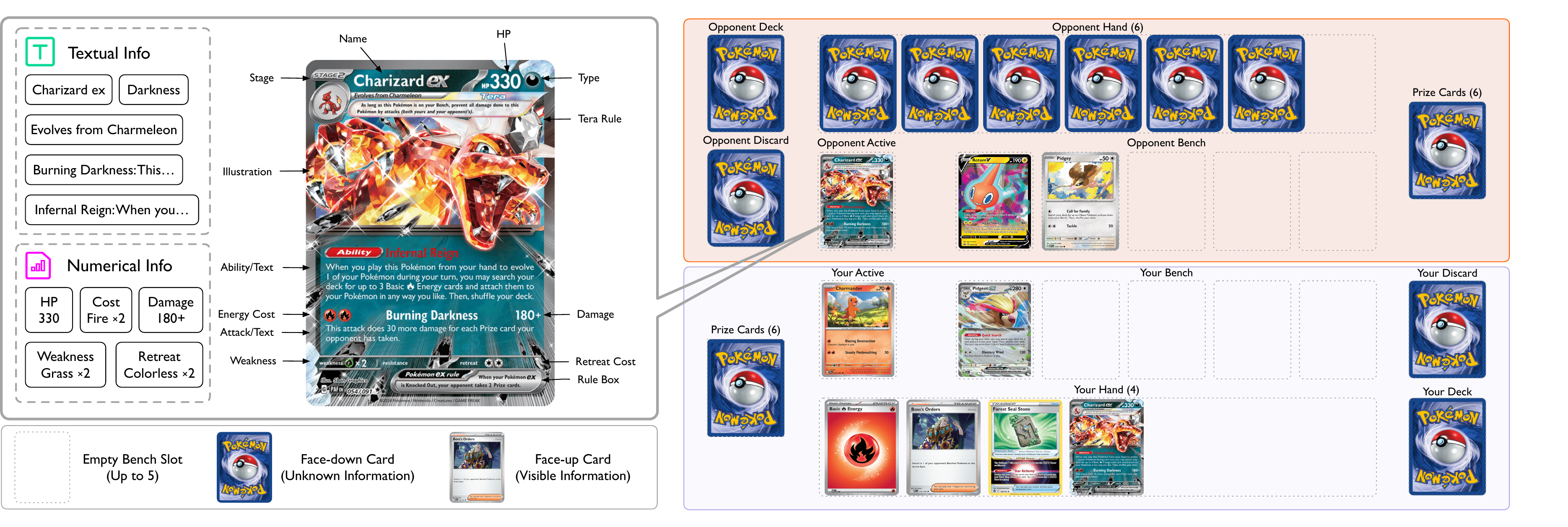}
    \caption{Overview of the environment of PTCG-Bench. \textbf{Left:} An example Pok\'{e}mon card annotated with key attributes including HP and Type, along with Ability and Attack descriptions that govern in-game decisions. \textbf{Right:} The two-player game board, where each player maintains an Active spot, up to five Bench slots, a Prize card zone, a Deck, a Discard pile, and a hand of hidden cards. Agents must reason over their own observable board state while inferring hidden information from the opponent's side.}
    \label{fig:overview}
\end{figure*}

\begin{table*}[t]
\centering
\small
\setlength{\tabcolsep}{5.5pt}
\renewcommand{\arraystretch}{1.12}
\caption{Comparison between PTCG-Bench and representative game-based benchmarks.
    Abbrev.: \textbf{IIR} = imperfect-information reasoning, \textbf{LHP} = long-horizon planning,
    \textbf{TNI} = text-numerical inference, \textbf{SDM} = strategic decision-making,
    \textbf{LSE} = long-term self-evolution, \textbf{MHS} = modular harness support,
    and \textbf{SGI} = single-game integration.
``\pmark'' denotes partial coverage, where the capability is present in the environment
or appears in some tasks but is not explicitly evaluated as a central requirement.}
\label{tab:game_bench_comparison}
\begin{tabular*}{\textwidth}{@{\extracolsep{\fill}}lccccccc@{}}
\toprule
\textbf{Benchmark}
& \textbf{IIR}
& \textbf{LHP}
& \textbf{TNI}
& \textbf{SDM}
& \textbf{LSE}
& \textbf{MHS}
& \textbf{SGI} \\
\midrule
ALE~\citep{bellemare2013arcade}
& \xmark & \pmark & \xmark & \pmark & \xmark & \xmark & \xmark \\
MiniGrid~\citep{chevalier2023minigrid}
& \pmark & \pmark & \xmark & \xmark & \xmark & \xmark & \xmark \\
Procgen~\citep{mohanty2021measuring}
& \xmark & \pmark & \xmark & \pmark & \xmark & \xmark & \xmark \\
TextWorld~\citep{cote2018textworld}
& \pmark & \pmark & \xmark & \pmark & \xmark & \xmark & \xmark \\
ALFWorld~\citep{shridhar2020alfworld}
& \pmark & \pmark & \pmark & \pmark & \xmark & \xmark & \xmark \\
MineDojo~\citep{fan2022minedojo}
& \pmark & \cmark & \pmark & \pmark & \pmark & \xmark & \cmark \\
\midrule
lmgame-Bench~\citep{hu2025lmgame}
& \pmark & \pmark & \pmark & \cmark & \xmark & \cmark & \xmark \\
Orak~\citep{park2025orak}
& \pmark & \pmark & \pmark & \cmark & \pmark & \cmark & \xmark \\
PokeAgent Challenge~\citep{karten2026pokeagent}
& \cmark & \cmark & \pmark & \cmark & \pmark & \pmark & \pmark \\
\midrule
\textbf{PTCG-Bench (Ours)}
& \cmark & \cmark & \cmark & \cmark & \cmark & \cmark & \cmark \\
\bottomrule
\end{tabular*}
\end{table*}

To address this gap, we introduce PTCG-Bench, a benchmark built upon the Pok\'{e}mon Trading Card Game (PTCG)\footnote{For an official introduction to the Pok\'{e}mon Trading Card Game, see \url{https://tcg.pokemon.com/en-us/learn/}.}, as shown in Figure~\ref{fig:overview}. PTCG is a strategic two-player zero-sum game in which agents must reason over partially observable game states, make decisions under stochastic card draws, integrate textual card effects with numerical attributes, and plan over adversarial interactions. This setting allows us to evaluate whether LLM agents can combine imperfect-information reasoning, long-horizon planning, hybrid textual-numerical inference, and strategic decision-making within a single coherent game, rather than solving them as isolated subtasks.

Beyond initial performance, PTCG-Bench evaluates a capability that is increasingly central to autonomous agents: whether they can improve through experience~\citep{ou2025symbolic}. For agents deployed in realistic environments, success cannot rely solely on static instructions or one-shot task solving; instead, agents must learn from trial and error, consolidate reusable experience, and adapt their future decisions as environments change. Recent self-improvement methods~\citep{agrawal2025gepa,xu2026mem,alzubi2026evoskill} and benchmarks~\citep{jimenez2024swe,zhou2024webarena,liu2024agentbench} have provided useful insights, but many still emphasize relatively static, short-horizon, or perfect-information settings. To better approximate the conditions faced by agents in realistic environments, PTCG-Bench evaluates self-evolving agents under a \textbf{longitudinal evaluation protocol}, in which self-evolving agents play a sequence of games, accumulate experience, and are evaluated over multiple rounds to determine whether such experience leads to improved performance in a dynamic environment.

PTCG-Bench further enables controlled analysis of \textbf{agent harness design}. Modern LLM agents depend not only on the backbone model, but also on the surrounding harness design~\citep{yang2024swe}. We therefore design a modular harness around three key factors: observation structure, legal-action masking, and context management. By ablating these components, PTCG-Bench allows us to quantify how harness choices affect gameplay performance and separate their effects from backbone capability.

We conduct extensive experiments and the results show that LLM agents can achieve competitive gameplay performance, but their performance varies widely across models and harness designs. PTCG-Bench produces a wide ranking of agent systems, with a 665-point Glicko-2 rating gap between the strongest and weakest LLM-based agents.\footnote{Under the Glicko-2 formula, a 665-point rating difference corresponds to an expected win probability of approximately 97.8\% for the higher-rated player.} Moreover, PTCG-Bench exposes the limitations of existing self-evolution methods, with most agents failing to improve consistently over time.

Our main contributions are as follows:
\begin{itemize}[nosep]
    \item We introduce \textbf{PTCG-Bench}, a PTCG-based benchmark for evaluating LLM agents in a complex environment involving imperfect information, long-horizon planning, and hybrid textual-numerical reasoning.

    \item We propose a \textbf{longitudinal evaluation protocol} to evaluate whether self-evolving agents can accumulate cross-game experience and convert it into improved subsequent decisions.

    \item We develop a \textbf{modular agent harness} with independently ablatable components, enabling controlled analysis of multiple harness designs beyond backbone capability.
\end{itemize}

\section{Preliminary}

\subsection{Pokémon Trading Card Game}

PTCG is a two-player, turn-based, zero-sum card game in which each player uses a 60-card deck composed mainly of Pok\'{e}mon, Energy, and Trainer cards. Players set up an Active Pok\'{e}mon, optional Benched Pok\'{e}mon, and six face-down Prize cards, then alternate turns to develop their board, manage resources, and attack the opponent. A player wins by taking all Prize cards, eliminating all opposing Pok\'{e}mon in play, or causing the opponent to draw from an empty deck.

PTCG is challenging for LLM agents because it combines hidden information, stochastic card draws, dynamic legal actions, and long-horizon resource planning. Agents must interpret natural-language card effects while reasoning over numerical quantities such as HP, damage, Energy costs, and Prize counts, and must make sequential decisions whose consequences can compound over many turns. These properties make PTCG a suitable environment for evaluating strategic decision-making and experience-based improvement. Complete rules are included in Appendix~\ref{sec:ptcg_rules}.

\subsection{LLM Agent Formalization}
\label{sec:formalization}

By absorbing the opponent policy into the environment dynamics, a PTCG match can be modeled as a partially observable Markov decision process (POMDP), where the complete game state includes hidden information such as the opponent's hand, face-down Prize cards, and unrevealed deck order. At step $t$, the agent receives an observation $o_t$, maintains an interaction history $h_t=(o_1,a_1,\ldots,o_{t-1},a_{t-1},o_t)$, selects an action $a_t$, and receives a reward $r_t$. Because the game state is not fully observable, the agent must act based on $o_t$ and $h_t$ rather than the complete state.

An LLM agent induces its policy through both a backbone model and a surrounding harness. The harness transforms the observation and history into a prompt, constrains or formats the available actions, and parses the model output into an executable action. We denote the resulting policy by
\begin{equation}
    a_t \sim \pi_{\theta,\eta}(\cdot \mid o_t, h_t),
\end{equation}
where $\theta$ represents the LLM backbone and $\eta$ represents the harness design. The expected performance of the agent can then be written as
\begin{equation}
    J(\theta,\eta)=\mathbb{E}_{\pi_{\theta,\eta}}\left[\sum_{t} r_t\right].
\end{equation}

\section{PTCG-Bench}

PTCG-Bench evaluates LLM-based agents in the Pok\'{e}mon Trading Card Game (PTCG) along three dimensions: \textbf{(i)} decision-making under imperfect information and long-horizon strategic interaction, \textbf{(ii)} self-evolution through accumulated cross-game experience, and \textbf{(iii)} the contribution of harness design beyond backbone capability. This section describes the game environment, the self-evolution protocol, the modular harness, the evaluation tournament, and the evaluation metrics.

\subsection{Benchmark Environment}
\label{sec:env}

\begin{figure*}[t]
  \centering
  \includegraphics[width=\linewidth]{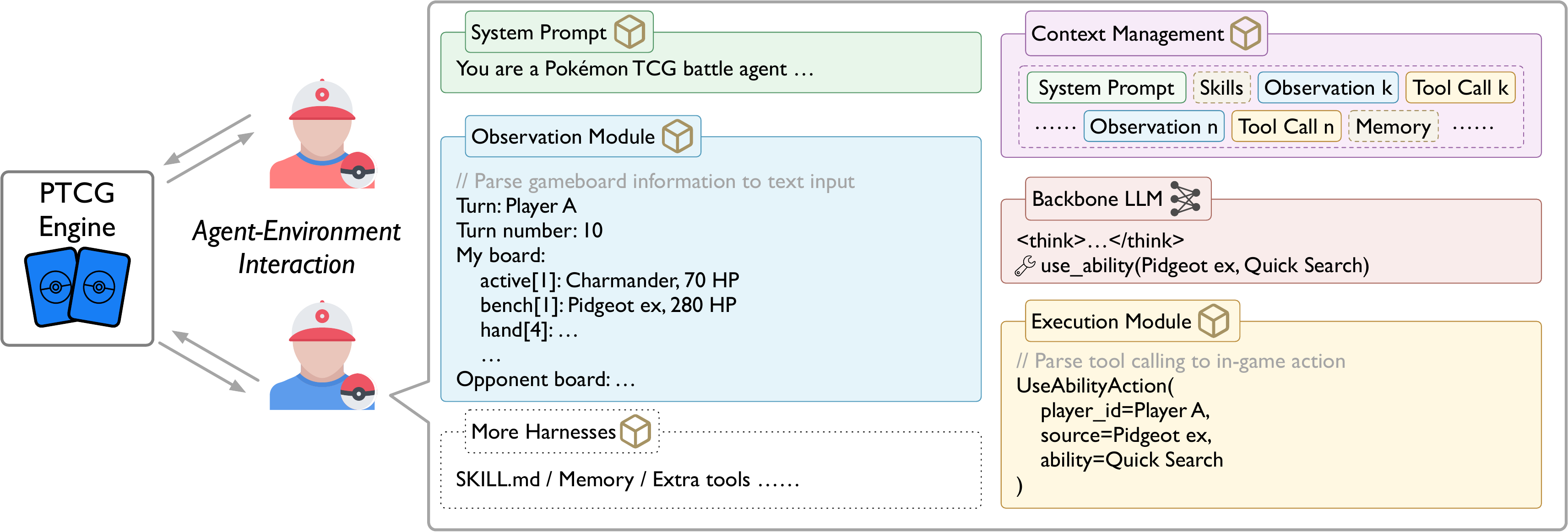}
  \caption{Agent-environment interaction loop in PTCG-Bench. At each decision point, the game engine exposes the current state and legal action set to the agent harness, which converts them into an LLM-readable prompt. The backbone model's response is parsed, validated, and executed by the engine to advance the game state.}
  \label{fig:agent_env}
\end{figure*}

\paragraph{Engine implementation.}
PTCG-Bench uses a custom PTCG engine\footnote{\url{https://github.com/gemelom/ptcg-engine}} for reproducible rule execution and legal-action validation. The engine implements the core mechanics needed for full-game evaluation, thereby providing a consistent environment for comparing agent systems. Detailed implementation details are provided in Appendix~\ref{sec:ptcg_rules}.

\paragraph{Deck pool.}
We use a fixed pool of 5 competitive decks covering distinct strategic archetypes, including aggressive, controlling, combo-oriented, and attrition-based plans. The deck pool is used both for controlled mirror-match evaluation and for cross-deck generalization analysis. Detailed deck statistics and strategic profiles are in Appendix~\ref{sec:deck_details}.

\paragraph{State and action spaces.}
PTCG-Bench separates private, public, and global state to preserve imperfect information. Agents observe their own private zones and public board information, while the opponent's hand, deck order, and Prize-card identities remain hidden. The action space is defined as a set of parameterized game actions exposed through tool schemas, including playing or evolving Pok\'{e}mon, attaching Energy, playing Trainer cards, attacking, retreating, and passing. Each action specifies the required arguments needed for execution, thereby providing a structured interface between agent decisions and game mechanics.

\subsection{Agent-Environment Interface}
\label{sec:interface}
As shown in Figure~\ref{fig:agent_env}, the engine exposes the game state at each decision point, from which the agent receives a partial observation. The agent harness combines this observation with the interaction history and renders them into an LLM-readable prompt. The backbone LLM then returns a structured action request, typically as a tool call with an action type and the corresponding parameters. The model response is parsed by the harness and checked by the engine against the current legal action set. Valid actions are executed to advance the game state, while invalid requests trigger a retry. This interface fixes the underlying game execution while allowing agent systems to vary in how they represent information, use history, and select actions. Such flexibility, in turn, enables evaluation with multiple harness modules.

\subsection{Longitudinal Evaluation Protocol}
\label{sec:protocol}

PTCG-Bench evaluates whether self-evolving agents improve through accumulated game experience. Games are organized into multiple rounds of complete PTCG matches. After each round, an evolving agent may update persistent state, such as reflections, strategic summaries, memory entries, revised prompts, or skill documents. This protocol separates within-game decision-making from cross-game experience updates.
We track round-wise performance against fixed-policy anchors to analyze whether accumulated experience improves, preserves, or degrades subsequent play. 

\begin{figure*}[t]
  \centering
  \includegraphics[width=\linewidth]{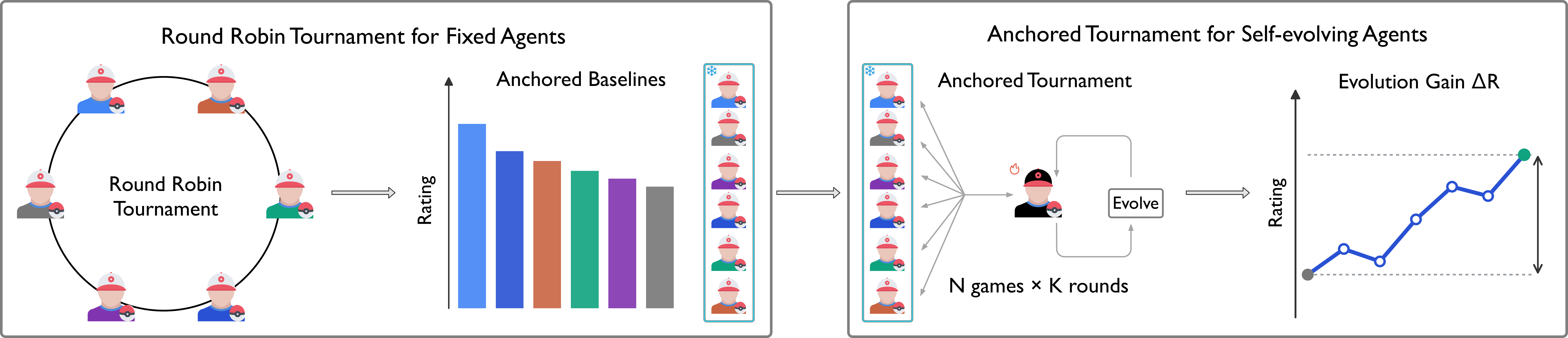}
  \caption{Evaluation tournament design in PTCG-Bench. \textbf{Left:} Fixed-policy agents are first evaluated by a round-robin tournament to obtain ratings, from which a set of fixed anchor baselines is selected. \textbf{Right:} Self-evolving agents are then evaluated through an anchored tournament, where each evolving policy plays against the frozen anchor baselines across rounds.}
  \label{fig:tournament}
\end{figure*}

\subsection{Modular Harness Architecture}
\label{sec:harness}
In PTCG-Bench, the harness is not treated as a neutral wrapper between the game engine and backbone LLM. Prior studies show that the surrounding agent scaffold, including state perception, context management, and action execution, can substantially change the measured capability of the same backbone model.

PTCG-Bench therefore decomposes the harness into modular components, allowing key agent-environment coupling factors to be ablated independently. This modular separation helps prevent the measured performance from conflating backbone capability with the effects of state perception, context management, and action execution. The specific harness modules and ablation settings are introduced in Section~\ref{sec:rq3}.

\subsection{Evaluation Tournament}
\label{sec:tournament}

As shown in Figure~\ref{fig:tournament}, PTCG-Bench uses two tournaments for different evaluation goals. Fixed agents are compared by round robin~\citep{harary1966theory}, where every configuration plays every other configuration to estimate overall playing strength on a shared rating scale. This setting supports direct comparison among static backbones and harness configurations.

Self-evolving agents use an anchored tournament because simultaneous policy updates would make a round-robin rating scale drift across rounds. Each evolving agent plays fixed-policy anchors with calibrated ratings spanning the rating range, then may update its persistent state. Anchor matchups therefore preserve a stable reference population across agent snapshots. Since anchors do not change, round-wise rating shifts primarily reflect changes in the evolving agent.

\subsection{Evaluation Metrics}
\label{sec:metrics}

PTCG-Bench reports 4 complementary metrics.

\paragraph{Glicko-2 rating.}
We use Glicko-2~\citep{glickman2012example} as the primary measure of playing strength, reporting both the rating mean $\mu$ and rating deviation $\phi$ under stochastic match outcomes. Ratings are estimated from round-robin records for fixed agents and from anchor matchups for evolving agents, with $\phi$ indicating estimation uncertainty.
    
\paragraph{Head-to-head win rate.}
Head-to-head win rates provide an interpretable complement to Glicko-2 by exposing matchup-level advantages that aggregate ratings can obscure.

\paragraph{Invalid action rate.}
Invalid action rate measures how often an agent produces illegal or unparsable actions before retry or fallback and is especially relevant to harness ablations.

\paragraph{Tool calls.}
The number of tool calls measures interaction cost, counting action attempts and operations invoked during decision-making. We use it to contextualize configuration efficiency.

\section{Experiments}

\subsection{Experimental Setup}
\label{sec:exp_setup}
In this section, we conduct comprehensive experiments to answer the following research questions: (\textbf{RQ1}) Can PTCG-Bench Resolve Strategic Capability Differences among LLM Agents? (\textbf{RQ2}) Can PTCG-Bench Measure Self-Evolution over Sequential Play? (\textbf{RQ3}) How Much Does Harness Design Affect Agent Performance?

\paragraph{LLM backbones.}
We evaluate 10 backbones from 5 model families, pairing a frontier and cost-efficient variant from each family: Gemini 3.1 Pro and Gemini 3 Flash, Claude Sonnet 4.6 and Claude Haiku 4.5, DeepSeek V4 Pro and DeepSeek V4 Flash, GPT-5.4 and GPT-5.4 Nano, and Qwen3.6 Plus and Qwen3.5 Flash. Unless otherwise specified, all agents use the same ReAct-style harness.

\paragraph{Evaluation configuration.}
Fixed-agent experiments use the round-robin tournament in Section~\ref{sec:tournament} with ten LLM agents and two non-LLM references, Random and Heuristic, which sample legal actions with uniform and predefined weights, respectively. Each pair plays $M=20$ games, yielding $1320$ games per tournament. Self-evolution uses the anchored tournament in Section~\ref{sec:tournament}. We compute ratings with Glicko-2 and use mirror matches unless stated otherwise.

\paragraph{Evaluation scale.}
We set $M=20$ to balance rating resolution and evaluation cost.
This scale provides sufficient resolution to distinguish agents with meaningful differences in strategic capability and produces stable anchor ratings for subsequent anchored evaluations.
At the same time, it keeps the evaluation computationally tractable, as each PTCG game involves long-horizon interaction and repeated model inference over many decision steps.

\begin{figure*}[t]
\centering
\includegraphics[width=\linewidth]{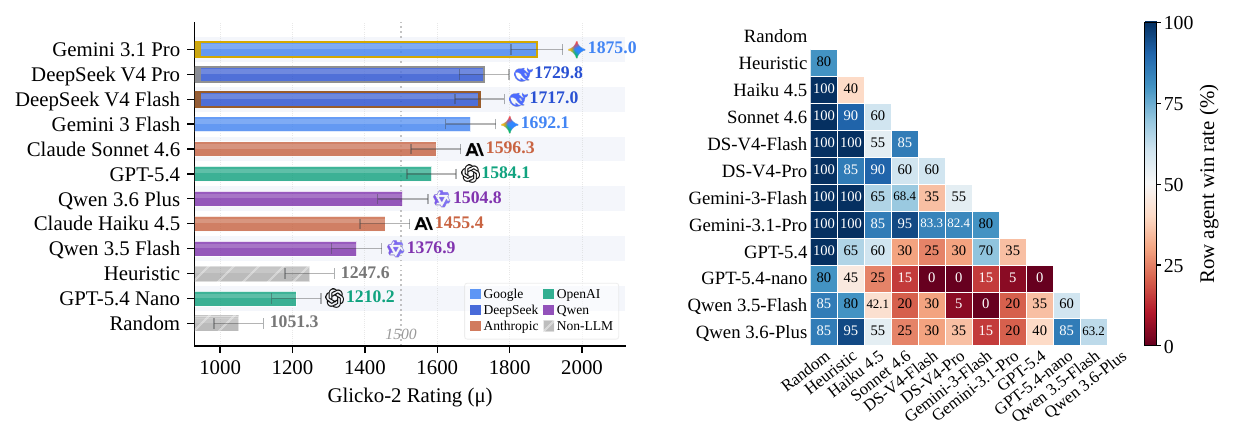}
\caption{PTCG-Bench tournament results under a unified ReAct harness. \textbf{Left:} Glicko-2 ratings ($\mu \pm \phi$) for LLM backbones and fixed-capability anchor agents, with all agents initialized at $\mu=1500$ and $\phi=350$; error bars indicate rating deviation. \textbf{Right:} pairwise win-rate heatmap, where each cell gives the row agent's win rate against the column agent. Higher ratings and colder heatmap values indicate stronger play.}
\label{fig:glicko_rq1}
\end{figure*}

\subsection{Backbone Strategic Capabilities}
\label{sec:rq1}

To answer \textbf{RQ1}, we test whether PTCG-Bench resolves backbone-level strategic differences under a fixed ReAct harness and mirror-match setting. Figure~\ref{fig:glicko_rq1} reports Glicko-2 ratings for all 10 backbones together with the fixed-capability anchors.

\textbf{PTCG-Bench produces a broad and well-resolved rating distribution.}
Among LLM-based agents, ratings span from GPT-5.4 Nano (1210) to Gemini 3.1 Pro (1875), corresponding to a gap of 665 Glicko-2 points. This spread substantially exceeds the rating deviations of most individual agents, suggesting that PTCG-Bench can separate LLM agents by playing strength rather than collapsing them into a narrow performance band. Head-to-head win rates in the pairwise heatmap further show that frontier variants consistently outperform their cost-efficient counterparts within the same model family, suggesting that PTCG-Bench captures meaningful capability differences rather than tournament noise.

\begin{figure}[t]
\centering
\includegraphics[width=\columnwidth]{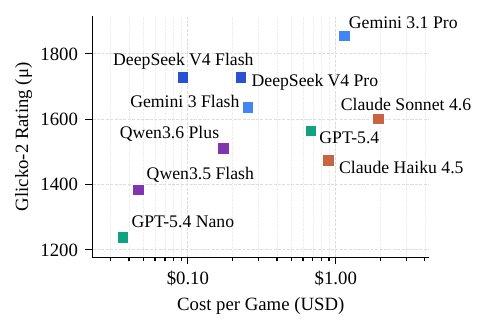}
\caption{Cost--rating trade-off for all ten LLM backbones. The $y$-axis reports the Glicko-2 rating mean $\mu$, and the $x$-axis reports inference cost per game in USD. Models closer to the upper-left region achieve stronger play at lower cost.}
\label{fig:cost_vs_rating}
\end{figure}

\textbf{Gameplay strength is not determined by inference cost.}
Figure~\ref{fig:cost_vs_rating} compares Glicko-2 rating with inference cost per game. The non-monotonic distribution shows that higher cost does not necessarily yield stronger play, suggesting that PTCG-Bench also reveals practical cost--performance trade-offs among LLM agents.

\textbf{PTCG-Bench captures capabilities not fully reflected by general-purpose benchmarks.}
We further compare the PTCG-Bench ranking with model orderings reported by other LLM evaluations such as LiveBench, SWE-Bench Pro, and GPQA~\citep{white2025livebench,deng2025swe,rein2023gpqa}. As detailed in Appendix~\ref{sec:external_rank_alignment}, the resulting rankings are only partially aligned. In particular, models that are stronger on general language understanding or preference-based evaluation do not always obtain proportionally higher PTCG-Bench ratings. This divergence suggests that PTCG-Bench evaluates capabilities beyond static language understanding, particularly imperfect-information reasoning, long-horizon planning, and strategic decision-making under uncertainty.

\subsection{Self-Evolution}
\label{sec:rq2}

\begin{figure*}[t]
\centering
\includegraphics[width=\textwidth]{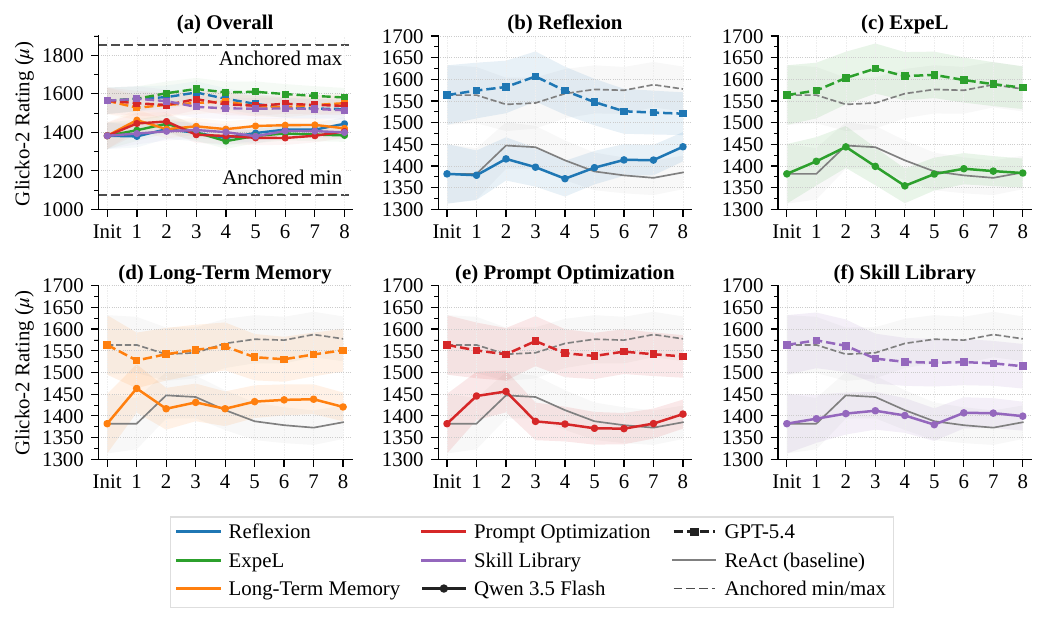}
\caption{
Glicko-2 rating trajectories across self-evolution rounds for five evolving agent configurations using Qwen3.5-Flash and GPT-5.4 as backbones.
Panel (a) summarizes all evolving configurations relative to the anchored rating range, with gray dashed lines denoting anchored maximum and minimum ratings and the dotted line denoting the non-evolving same-backbone reference.
Panels (b)--(f) provide method-specific comparisons against the non-evolving ReAct baseline, with shaded regions indicating rating uncertainty bands.
}
\label{fig:learning_curve}
\end{figure*}

To answer \textbf{RQ2}, we evaluate whether PTCG-Bench can track \emph{self-evolution} across sequential play using fixed anchors and a stable rating scale.

\paragraph{Evolving agent configurations.}
We evaluate five representative self-evolving baselines under the same PTCG-Bench harness: Reflexion~\citep{shinn2023reflexion}, ExpeL~\citep{zhao2024expel}, long-term memory~\citep{park2023generative,packer2023memgpt,tan2025prospect,xu2026mem}, prompt evolution~\citep{madaan2023self,wang2024promptagent,yuksekgonul2025optimizing}, and skill-library evolution~\citep{agentskills2026,alzubi2026evoskill}. Since PTCG involves long horizons, delayed high-variance outcomes, hidden information, stochastic card order, and opponent-dependent strategy shifts, directly transplanting rollout- or search-based methods would be costly and unstable. We therefore adapt these baselines at the mechanism level. All configurations use the same backbone and harness to isolate the effect of self-evolution; details are provided in Appendix~\ref{sec:self_evolution_details}.

\paragraph{Anchored self-evolution protocol.}
We instantiate each of the five self-evolution mechanisms with two backbones, Qwen3.5-Flash and GPT-5.4, and evaluate each resulting configuration for $R=8$ rounds. In each round, every configuration plays 2 games against each of 12 fixed anchors: the 10 static LLM agents from RQ1 plus the Random and Heuristic agents.
For each backbone, we additionally report the corresponding non-evolving ReAct agent as a static reference. This yields 24 games per configuration, $N=120$ evolving-agent games per backbone per round, and 240 games per round across both backbones. After each round, each mechanism updates its persistent state using the accumulated trajectories before the next snapshot is evaluated against the same anchors, keeping rating changes on a stable reference scale.

\begin{table*}[t]
\centering
\small
\setlength{\tabcolsep}{4.2pt}
\caption{Main harness ablation results. All configurations use the same LLM backbone and mirror-match tournament setting. $\mu$ denotes the Glicko-2 rating mean, and $\Delta\mu$ is measured relative to the full harness. Values in parentheses normalize the rating drop by the median rating gap between adjacent LLM backbones in the RQ1 ranking ($55$ rating points), which is also reported in the final row as a reference scale. Inv. Rate denotes the fraction of illegal or unparsable actions before retry or fallback.}
\label{tab:harness_ablation}
\begin{tabular*}{\textwidth}{@{\extracolsep{\fill}}lccccccc@{}}
\toprule
\textbf{Configuration} 
& \textbf{Struct.} 
& \textbf{Mask.} 
& \textbf{Hist.} 
& \textbf{$\mu$} 
& \textbf{$\Delta\mu$}
& \textbf{Inv. Rate} 
& \textbf{Tool Calls} \\
\midrule
Full Harness (\ref{item:obs-structure}+\ref{item:legal-mask}+\ref{item:context-history})
& \cmark & \cmark & \cmark
& 1726
& --
& 3.3\%
& 78.8 \\

w/o Structured Observation (w/o \ref{item:obs-structure})
& \xmark & \cmark & \cmark
& 1693
& $-33$ ($0.6\times$)
& 5.2\%
& 68.4 \\

w/o Legal Action Masking (w/o \ref{item:legal-mask})
& \cmark & \xmark & \cmark
& 1608
& $-118$ ($2.1\times$)
& 15.9\%
& 86.7 \\

No History Context (w/o \ref{item:context-history})
& \cmark & \cmark & \xmark
& 1611
& $-115$ ($2.1\times$)
& 19.0\%
& 385.7 \\

Minimal Harness (w/o \ref{item:obs-structure}+\ref{item:legal-mask}+\ref{item:context-history})
& \xmark & \xmark & \xmark
& 1575
& $-151$ ($2.7\times$)
& 27.3\%
& 357.6 \\
\midrule
Median adjacent LLM gap
& \cmark & \cmark & \cmark
& 55
& $1.0\times$
& \multicolumn{2}{c}{--} \\
\bottomrule
\end{tabular*}
\end{table*}

\subsection{Modular Harness Ablation}
\label{sec:rq3}

Figure~\ref{fig:learning_curve} reports the resulting rating trajectories and compares them with anchored baseline ratings and the non-evolving same-backbone ReAct reference. The anchored rating range suggests substantial headroom for current self-evolving baselines, yet none of the five mechanisms shows consistent monotonic improvement across the eight rounds. Ratings instead fluctuate without forming a stable upward trend, and no evolving configuration reliably surpasses the non-evolving same-backbone reference by the end of evaluation. These results suggest that PTCG-Bench reveals the incompleteness of current self-evolution mechanisms: existing agents still struggle to extract reusable strategic knowledge from long-horizon, imperfect-information gameplay without dense rewards or verifiable ground-truth solutions.

To answer RQ3, we examine whether and to what extent harness design affects measured agent performance. We hold the LLM backbone, deck pool, and tournament setting fixed, and ablate key harness choices that govern state representation, action execution, and context management.

\paragraph{Harness modules.}
We focus on these three modules because they correspond to the main stages through which the harness can influence gameplay: how the agent perceives the state, how it grounds decisions into executable actions, and how it retains temporal information across turns.
\begin{enumerate}[nosep,leftmargin=15pt,label=\textbullet,ref=H\arabic*]
    \item \label{item:obs-structure}
    \textbf{Observation structure (H1)}: replacing the structured state description with a minimally formatted raw state or game-log representation.
    
    \item \label{item:legal-mask}
    \textbf{Legal-action masking (H2)}: removing the engine-computed legal action set and requiring the agent to infer valid actions and parameters from the current state.
    
    \item \label{item:context-history}
    \textbf{Context management (H3)}: removing recent trajectory history while retaining the current state and legal actions.
\end{enumerate}

\paragraph{Main harness ablation.}
Table~\ref{tab:harness_ablation} reports the main harness ablation results. The full harness uses structured observation, legal-action masking, and context-window-aware history truncation. We compare it against three single-component ablations and a minimal harness. The minimal harness removes all three interface supports: it uses unstructured observation, disables legal-action masking, and provides no recent history.

The ablations show that harness design affects measured capability. Removing structured observation yields a moderate 33-point rating drop and more invalid actions, whereas removing legal-action masking or recent history reduces rating by 118 and 115 points, respectively; the former increases invalid rate, and the latter increases tool calls by nearly fivefold. These drops exceed typical adjacent-backbone gaps in RQ1 and approach several within-family model-tier gaps. Harness design therefore has a substantial effect on measured gameplay strength, making it a central factor in evaluating LLM-agent systems with PTCG-Bench.

\subsection{Cross-Deck Generalization}
\label{sec:cross_deck}

We further evaluate deck-wise mirror-match generalization across five deck archetypes to test whether the measured backbone differences persist under different strategic settings. The main agent ranking is largely preserved across decks, suggesting that the measured differences are not artifacts of a single mirror deck. Details are in Appendix~\ref{sec:cross_deck_generalization}.

\section{Related Work}

\subsection{Self-Evolving LLM Agents}

Recent work has explored self-evolving LLM agents that improve through interaction histories, reflective feedback, and reusable experience. GEPA evolves prompts from reflective trajectory analysis~\citep{agrawal2025gepa}; A-MEM organizes adaptive memories for long-term improvement~\citep{xu2026mem}; and EvoSkill refines reusable skills from execution failures~\citep{alzubi2026evoskill}. These studies suggest that agents can improve not only through model-level training~\citep{da2025agent,huang2026policysim,zhang2026rapo}, but also through evolving prompts, memories, and skills.

\subsection{Benchmarks for LLM Game Agents}

Games are important testbeds because they require perception, memory, planning, and long-horizon decision-making. LMGame-Bench~\citep{hu2025lmgame} evaluates diverse games with modular perception, memory, and reasoning components, while Orak~\citep{park2025orak} covers multiple video-game genres for systematic agent training and evaluation. Other benchmarks emphasize specific challenges: TCG-Bench~\citep{alrashed2026cards} is a contamination-resistant and difficulty-scalable multilingual card-game benchmark, while the PokeAgent Challenge~\citep{karten2026pokeagent} targets competitive decision-making under partial observability and long-context strategic reasoning.

\section{Conclusion}

We introduced \textbf{PTCG-Bench}, a comprehensive benchmark for studying LLM agents in a realistic and complex game environment. By combining long-horizon imperfect-information gameplay with a longitudinal evaluation protocol, PTCG-Bench provides a controlled setting for evaluating self-evolution under delayed feedback, stochasticity, and opponent interaction. Experiments across ten LLM backbones and multiple self-evolution mechanisms show that current agents can achieve meaningful gameplay performance, but accumulated experience does not yet reliably translate into stable performance gains. Harness ablations further show that interface design materially affects measured capability, underscoring the need to study self-evolution at the agent-system level. These findings highlight PTCG as a challenging testbed for future research on self-evolving agents.

\section*{Limitations}

PTCG-Bench is designed to support controlled evaluation of LLM-agent systems in a complex trading card game, and the current study focuses on settings that make performance differences interpretable. In particular, we use a fixed deck pool and primarily rely on mirror-match evaluation to reduce asymmetric matchup effects, while studying experience-based updates to prompts, memories, and skills under a fixed LLM backbone. These choices allow us to isolate backbone capability, harness design, and self-evolution mechanisms within the same environment, but they do not exhaust the full space of strategic adaptation. Future extensions could incorporate open deck building, arbitrary cross-deck matchups, longer experience streams, or parameter-level learning while retaining the same controlled evaluation framework.

\section*{Ethical Considerations}

In constructing PTCG-Bench, we utilized publicly available information from the Pokémon Trading Card Game (PTCG), including card names, mechanics, descriptions and images. All materials are used solely for non-commercial academic research purposes under the principles of fair use / research exceptions in applicable copyright laws.
This benchmark is not intended for commercial applications, nor does it claim any affiliation with, sponsorship by, or endorsement from The Pokémon Company, Nintendo, or related entities. We respect the intellectual property rights of the copyright holders and have made efforts to use the data in a transformative manner for benchmarking language models. The dataset will be released only for research use, and we encourage users to comply with all relevant IP regulations.

The data used in PTCG-Bench is derived from fictional card game materials and does not involve human subjects, user-generated personal data, or information that uniquely identifies real individuals. We reviewed the collected data to ensure that it does not contain personally identifying information.

We used ChatGPT to assist with language editing and improving the clarity of the manuscript. The authors reviewed and verified all AI-assisted content and are responsible for the final paper.

\section*{Acknowledgments}
This work is supported by NSFC (No.\ 62206056, No.\ 62322606, No.\ 62441605, No.\ 62606466), the CIPSC-SMP-Zhipu Large Model Cross-Disciplinary Fund, and a collaboration funding by FinVolution Group.

\bibliography{references}

\appendix
\appendix

\appendix

\section{PTCG Rules and Engine Details}
\label{sec:ptcg_rules}

\subsection{Complete {PTCG} Rules}
\label{sec:complete_ptcg_rules}

PTCG-Bench follows the official rules of the Pok\'{e}mon Trading Card Game. Each game is played between two players with 60-card decks, hidden hands and decks, public boards, discard piles, and six Prize cards. A player first sets up an Active Pok\'{e}mon and optional Benched Pok\'{e}mon, then alternates turns consisting of card draw, legal in-turn actions, and optional attack resolution. During a turn, players may play Basic Pok\'{e}mon, evolve eligible Pok\'{e}mon, attach Energy, play Trainer cards, use Abilities, retreat or switch the Active Pok\'{e}mon, and attack when the required conditions are satisfied. The engine implements the main card-type mechanics for Pok\'{e}mon, Energy, Items, Supporters, Tools, and Stadiums, as well as evolution constraints, retreat costs, Special Conditions, Knock Outs, Prize-taking, and standard win conditions. A player wins by taking all Prize cards, by leaving the opponent with no Pok\'{e}mon in play, or when the opponent cannot draw at the beginning of their turn. For complete official rules and card-specific timing details, we refer readers to the official PTCG rulebook~\citep{pokemon-tcg-rulebook}.

\subsection{Engine Implementation}
\label{sec:engine_implementation}

The PTCG-Bench engine is implemented in Python to facilitate integration with common agent research pipelines. It maintains the full game state, validates legal actions before execution, applies rule-based state transitions, and records complete game trajectories for later analysis. This unified interface allows agents with different prompting, planning, memory, or learning mechanisms to interact with the same environment under reproducible execution.

The implemented card metadata, attacks, abilities, and rule interactions follow the official Pokémon TCG card texts and game rules. Engine correctness is validated through a comprehensive \texttt{pytest} suite with approximately 95\% code coverage. At the card level, every implemented card is systematically tested for metadata, attack effects, abilities, zone transitions, player-choice flows, damage and Knock Out resolution, and negative or edge-case behavior, such as invalid targets and unmet activation conditions. At the deck level, domain experts conducted complete matches using all five benchmark decks to identify implementation inconsistencies and validate complex cross-card interactions. These automated and expert-driven validations jointly ensure that the engine preserves the gameplay mechanics and strategic interactions required by the benchmark.

\begin{figure}[t]
    \centering
    \includegraphics[width=\linewidth]{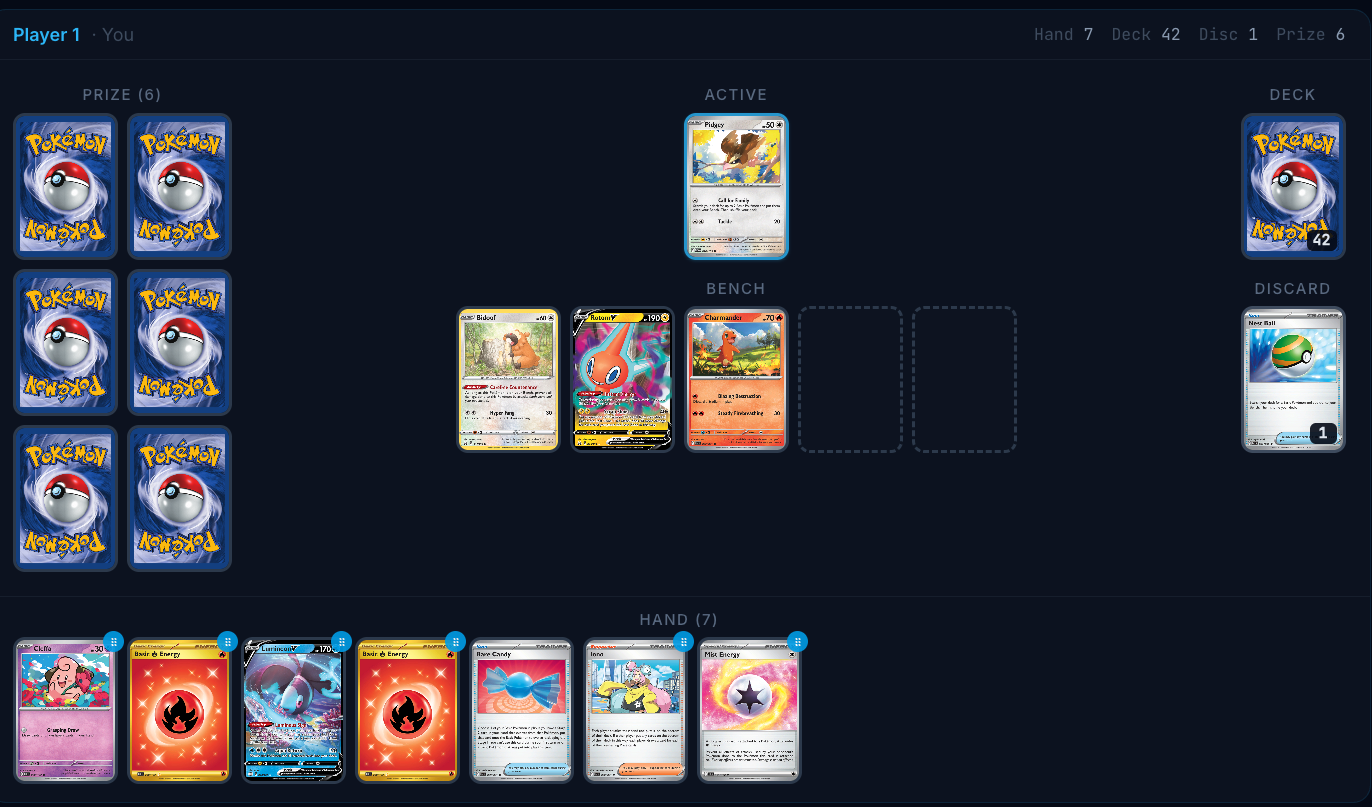}
        \caption{Screenshot of the PTCG-Bench frontend. The interface visualizes our half-field board state, including the Active Pok\'{e}mon, Bench, Prize cards, Deck, Discard pile, and Hand.}
    \label{fig:engine_ui}
\end{figure}

As shown in Figure~\ref{fig:engine_ui}, we also provide a React-based frontend for replay, debugging, and qualitative inspection. The frontend visualizes public game states, action histories, card movements, Prize-card progression, and agent decisions over time, making it easier to trace failure cases and interpret agent behavior during full-game evaluation. The frontend is used only for visualization and analysis; game execution and scoring are determined by the Python engine.

\section{Deck Details}
\label{sec:deck_details}

PTCG-Bench uses a fixed pool of five competitive decks to cover representative strategic archetypes while keeping the evaluation space controlled. Each deck induces a different pattern of board development, resource management, and win-condition execution, requiring agents to adapt across heterogeneous strategic contexts. Table~\ref{tab:decks} summarizes the composition and strategic profile of each deck.

\begin{table*}[t]
\centering
\small
\setlength{\tabcolsep}{3.5pt}
\caption{Deck composition and strategic profiles of the five PTCG-Bench decks. \#P/\#T/\#E denote Pok\'{e}mon, Trainer, and Energy card counts, respectively.}
\label{tab:decks}
\begin{tabular}{lccc p{1.7cm} p{2.3cm} p{3.0cm} p{3.6cm}}
\toprule
\textbf{Deck} 
& \textbf{\#P} 
& \textbf{\#T} 
& \textbf{\#E} 
& \textbf{Plan} 
& \textbf{Board Structure} 
& \textbf{Resource Focus} 
& \textbf{Agent Challenge} \\
\midrule
Charizard ex   
& 20 & 32 & 8  
& Aggro / scaling 
& Stage-2 evolution line 
& Rare Candy setup; Fire Energy acceleration 
& Timing evolution and prize-based damage scaling \\

Gardevoir ex   
& 15 & 34 & 11 
& Control / recursion 
& Stage-2 evolution line 
& Psychic Energy recovery; discard management 
& Balancing damage, Energy recursion, and attacker rotation \\

Miraidon ex    
& 13 & 29 & 18 
& Fast aggro 
& Basic-Pok\'{e}mon axis 
& Electric Energy acceleration; early board filling 
& Sequencing search, acceleration, and early pressure \\

Gholdengo ex   
& 19 & 30 & 11 
& Combo / burst 
& Multi-attacker box 
& Hand size; Energy discard and recovery 
& Estimating burst damage and preserving combo resources \\

Lugia Archeops 
& 20 & 24 & 16 
& Attrition / toolbox 
& Multi-core box 
& Special Energy acceleration; Archeops setup 
& Choosing attackers and allocating Special Energy efficiently \\
\bottomrule
\end{tabular}
\end{table*}

Together, these decks contain over 100 unique cards and span several axes of strategic diversity, including aggressive versus controlling plans, fast setup versus resource-efficient play, single-core versus multi-core win conditions, and evolution-line versus Basic-Pok\'{e}mon-centered board construction. This deck pool reduces the risk that evaluation results are dominated by a single play pattern, while remaining small enough to support controlled mirror-match and cross-deck analyses.

\section{Agent-Environment Interface Details}
\label{sec:interface_details}

This appendix provides additional details on the agent-environment interface used in PTCG-Bench. All LLM agents interact with the game engine through the same structured observation format and tool-based action interface. At each decision step, the engine provides the current public and private game state, the legal action set, and any pending card-selection prompt. The field \texttt{available\_actions} is treated as authoritative: agents are instructed to select only actions listed by the engine. A game turn may therefore consist of multiple LLM decision steps, since each tool call executes a single game action and returns an updated observation.

\paragraph{Agent Configuration.}
Unless otherwise specified, all agents use the same decoding and interaction settings within each experiment. We fix the system prompt template, observation format, action schema, retry/fallback policy, and context-management rule across compared agents. The default harness maintains recent trajectory context up to the model's maximum context window and truncates older interaction history when necessary. The no-history ablation removes trajectory history entirely while preserving the current state and legal action set. For self-evolving agents, the LLM backbone and action interface are held fixed across rounds; only the persistent state associated with the evolution mechanism, such as reflections, lessons, memories, revised prompts, or skills, is updated.

\paragraph{Tool Interface.}
The tool interface separates executable game actions from information-query tools. Game-action tools modify the game state and must correspond to legal actions returned by the engine. Query tools do not modify the game state and are used to inspect card text or discard-pile contents. The \texttt{activate\_skill} tool is enabled only for the skill-library configuration and retrieves stored procedural guidance relevant to the current state. Table~\ref{tab:tool_schema} summarizes the tool schema exposed to agents.

\begin{table*}[t]
\centering
\small
\setlength{\tabcolsep}{4pt}
\caption{Summary of the PTCG-Bench tool interface. Required arguments are shown in parentheses; optional disambiguation indices are omitted for brevity.}
\label{tab:tool_schema}
\begin{tabular*}{\textwidth}{@{\extracolsep{\fill}}lll@{}}
\toprule
\textbf{Category} & \textbf{Tool} & \textbf{Purpose and Required Arguments} \\
\midrule
Attack 
& \texttt{attack} 
& Use the Active Pok\'{e}mon's attack; requires \texttt{source\_card}, \texttt{attack\_name}. \\

Board setup 
& \texttt{play\_pokemon} 
& Play a Basic Pok\'{e}mon to the Active Spot or Bench; requires \texttt{source\_card}, \texttt{position}. \\

Evolution 
& \texttt{evolve\_pokemon} 
& Evolve a Pok\'{e}mon in play; requires \texttt{source\_card}, \texttt{target\_card}. \\

Energy 
& \texttt{attach\_energy} 
& Attach one Energy card from hand; requires \texttt{source\_card}, \texttt{target\_card}. \\

Trainer 
& \texttt{use\_supporter} 
& Play a Supporter card; requires \texttt{source\_card}. \\

Trainer 
& \texttt{use\_item} 
& Play an Item card; requires \texttt{source\_card}. \\

Trainer 
& \texttt{use\_tool} 
& Attach a Pok\'{e}mon Tool; requires \texttt{source\_card}, \texttt{target\_card}. \\

Stadium 
& \texttt{put\_stadium} 
& Play a Stadium card; requires \texttt{source\_card}. \\

Stadium 
& \texttt{discard\_stadium} 
& Discard the current Stadium when legally allowed; requires \texttt{source\_card}. \\

Stadium 
& \texttt{use\_stadium} 
& Activate an in-play Stadium effect; requires \texttt{source\_card}. \\

Ability 
& \texttt{use\_ability} 
& Activate a Pok\'{e}mon Ability; requires \texttt{source\_card}, optionally \texttt{ability\_name}. \\

Switching 
& \texttt{retreat} 
& Retreat the Active Pok\'{e}mon; requires \texttt{source\_card}. \\

Selection 
& \texttt{choose\_card} 
& Resolve a card-selection prompt; requires \texttt{chosen\_cards}. \\

Turn control 
& \texttt{pass\_turn} 
& End the current turn without further action. \\

Information 
& \texttt{query\_card} 
& Query exact card text and metadata; requires \texttt{card\_id}. \\

Information 
& \texttt{query\_discard} 
& Inspect a player's discard pile; requires \texttt{player}. \\

Skill retrieval 
& \texttt{activate\_skill} 
& Load stored skill guidance; requires \texttt{name}, optionally \texttt{resource}. \\
\bottomrule
\end{tabular*}
\end{table*}

For actions involving duplicate Pok\'{e}mon with the same name, the observation may provide field indices to disambiguate targets. Agents are instructed to copy card names, attack names, target names, and indices exactly from the observation or \texttt{available\_actions}. Invalid, unparsable, or unavailable actions are rejected by the game engine and counted in the invalid action rate before retry or fallback.

\begin{figure*}[t]
\centering
\includegraphics[width=\textwidth]{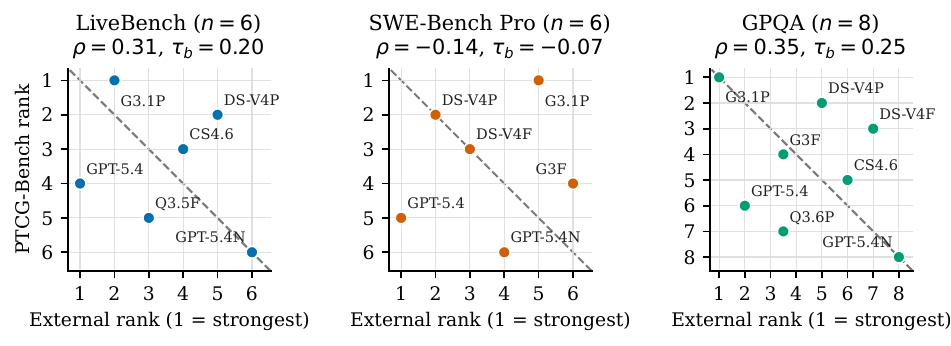}
\caption{Rank agreement between PTCG-Bench and external LLM evaluations on available overlapping model subsets. Each point is one shared model, and the dashed diagonal marks identical orderings within the overlap subset. LiveBench and GPQA show weak positive agreement with PTCG-Bench, while the collected SWE-Bench Pro ordering shows no clear agreement.}
\label{fig:external_rank_alignment}
\end{figure*}

\section{Self-Evolution Mechanism Details}
\label{sec:self_evolution_details}

We compare self-evolution mechanisms at the level of how prior gameplay experience is converted into persistent decision-making state. This design keeps the PTCG interaction interface fixed while varying the form of experience retained across games. In particular, all evolving configurations share the same backbone model, base harness, action tools, and evaluation anchors; they differ only in whether past trajectories are converted into reflections, distilled lessons, retrieved memories, revised prompt instructions, or structured skills.

PTCG requires mechanism-level adaptation rather than direct reuse of procedures designed for short-horizon or verifiable tasks. A complete trajectory contains many tool-mediated decisions, hidden opponent information, stochastic card draws, and a delayed game outcome. We therefore use completed games and round-level trajectory collections as the experience source for persistent updates, while later matches remain the evaluation signal for whether those updates transfer to new gameplay.

\paragraph{Reflexion.}
Following reflection-based verbal reinforcement~\citep{shinn2023reflexion}, the agent produces a free-form reflection after each game. The reflection summarizes mistakes, missed opportunities, and strategic lessons inferred from the trajectory and terminal outcome. These reflections remain episodic: they preserve game-specific feedback and are prepended to later game contexts so the agent can condition subsequent decisions on recent self-critique. This configuration tests whether lightweight post-game reflection alone can convert individual failures into better future play.

\paragraph{ExpeL.}
ExpeL-style evolution~\citep{zhao2024expel} aggregates experience before updating persistent state. After each evaluation round, the agent distills accumulated trajectories into reusable strategic lessons rather than storing one reflection per game. The lessons aim to abstract recurring patterns across games, such as common planning failures or generally useful decision rules, and are provided to later games as compact guidance. This configuration separates cross-game lesson distillation from the more episodic feedback used by Reflexion.

\paragraph{Long-Term Memory.}
The long-term memory configuration follows memory-based agent designs~\citep{park2023generative,packer2023memgpt,tan2025prospect,xu2026mem}. It stores prior gameplay episodes externally and retrieves memories judged relevant to the current game state during later play. To reduce accumulation of low-level trajectory detail, periodic summarization compresses past episodes into higher-level strategic notes while preserving access to experience-derived guidance. This configuration tests whether retrieval over prior experience is more useful than injecting a fixed set of lessons into every later context.

\paragraph{Prompt Evolution.}
Prompt evolution treats the strategy prompt itself as the persistent state~\citep{madaan2023self,wang2024promptagent,yuksekgonul2025optimizing}. After each evaluation round, the agent analyzes the current strategy prompt together with recent game summaries, identifies recurring failure modes, and revises the strategy component of its instruction prompt. The revised prompt is then used in subsequent rounds under the same interaction interface. Unlike memory-based variants, this mechanism encodes experience into global decision principles that shape all later decisions rather than retrieving state-specific past episodes.

\paragraph{Skill Library Evolution.}
Skill-library evolution maintains a persistent collection of structured strategy skills~\citep{agentskills2026,alzubi2026evoskill}. Skills are distilled from prior games and describe an activation condition, a strategic objective, and recommended decision principles. During later play, relevant skills are retrieved when their guidance matches the current decision context. This configuration provides a more modular form of persistent knowledge than a monolithic revised prompt: different reusable skills can target different strategic situations while remaining grounded in accumulated gameplay experience.

Across these configurations, the persistent state is updated only by the corresponding self-evolution mechanism. The environment state, legal-action interface, and fixed anchor opponents are unchanged across rounds, allowing the longitudinal results in Section~\ref{sec:rq2} to compare how different forms of retained experience affect subsequent gameplay strength.

\section{External Benchmark Rank Agreement}
\label{sec:external_rank_alignment}

We compare the PTCG-Bench backbone ordering with model scores reported by LiveBench, SWE-Bench Pro, and GPQA~\citep{white2025livebench,deng2025swe,rein2023gpqa}. Each comparison is computed only on the subset of evaluated PTCG-Bench backbones with an available score for that external benchmark. We rerank PTCG-Bench and the external benchmark within that overlap subset, with rank 1 denoting the strongest model, and report both Spearman $\rho$ and Kendall $\tau_b$ rank correlations in Figure~\ref{fig:external_rank_alignment}. Tied external scores receive average ranks; this affects the two GPQA entries with score 90.4.

On the overlapped model subsets, LiveBench yields Spearman $\rho=0.31$ and Kendall $\tau_b=0.20$, GPQA yields $\rho=0.35$ and $\tau_b=0.25$, and SWE-Bench Pro yields $\rho=-0.14$ and $\tau_b=-0.07$. This suggests that PTCG-Bench measures critical aspects of strategic reasoning in complex, long-horizon game scenarios that are insufficiently captured by these existing benchmarks.

\section{Cross-Deck Generalization}
\label{sec:cross_deck_generalization}

\begin{figure}[t]
    \centering
    \includegraphics[width=\linewidth]{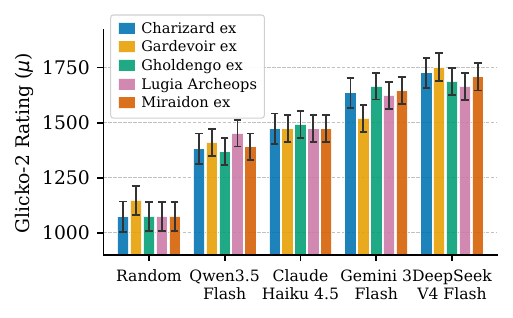}
    \caption{Cross-deck mirror-match generalization results. Each group of bars represents agents evaluated under one deck archetype, with both players using the same deck within that setting. Bar height is the Glicko-2 rating $\mu$, and error bars denote rating deviation $\phi$.}
    \label{fig:cross_deck}
\end{figure}

We test cross-deck generalization with deck-wise mirror matches, varying the shared deck archetype across settings while avoiding asymmetric matchup effects. Figure~\ref{fig:cross_deck} reports results across five representative decks: Charizard ex, Gardevoir ex, Miraidon ex, Gholdengo ex, and Lugia Archeops.

The RQ1 ranking is largely preserved across all five deck archetypes. Although individual ratings fluctuate with deck mechanics, these variations do not produce systematic rank reversals, indicating that the measured backbone differences are not artifacts of a single mirror deck. This controlled analysis tests cross-archetype robustness rather than asymmetric matchups between different decks.

\section{Prompt Templates}
\label{sec:prompts}

The following prompt is used for the non-evolving ReAct agent and as the base interaction template for evolving variants. The full prompt additionally includes concise reminders of PTCG-specific rules, including evolution restrictions, first-turn restrictions, retreat constraints, Trainer-card limits, damage calculation, Special Conditions, and Knock Out resolution.

\begin{quote}
\small
\textbf{Role.}
You are a Pok\'{e}mon TCG battle agent using the ReAct pattern. At each decision step, analyze the current game state and then call exactly one tool.

\textbf{Objective.}
Win the game by taking all Prize cards, Knocking Out all of the opponent's Pok\'{e}mon in play, or causing the opponent to be unable to draw at the beginning of their turn.

\textbf{Turn Structure.}
Each turn consists of drawing a card, taking legal in-turn actions, and optionally attacking. Legal in-turn actions include playing Basic Pok\'{e}mon, evolving eligible Pok\'{e}mon, attaching one Energy card, playing Trainer cards, using Abilities, retreating the Active Pok\'{e}mon, and passing the turn. Attacking ends the turn after damage, effects, Knock Outs, Prize-taking, and win checks are resolved.

\textbf{Observation Reading.}
The current game state is provided as a structured observation. The field \texttt{available\_actions} is authoritative: choose only actions listed there. If \texttt{choosing\_card} is true, respond to that selection prompt before taking any other action. Use \texttt{opponent\_last\_turn\_actions} to infer recent changes. Your hand is visible, while the opponent's hand and deck remain hidden.

\textbf{Decision Pattern.}
Before acting, consider HP and damage, Energy attachments, available attacks, Prize counts, immediate threats, and possible win conditions. Then call the appropriate tool. A full game turn may require multiple decision steps because each game action is executed separately.

\textbf{Action Discipline.}
Call at most one game-action tool in each response. Use exact card names, attack names, targets, and indices copied from the observation or \texttt{available\_actions}. Do not invent actions, targets, card names, or attack names. If an intended action is not listed in \texttt{available\_actions}, it is not currently legal. If an action fails or produces an unexpected result, inspect the updated observation and query relevant cards before proceeding.

\textbf{Tools.}
Use query tools, such as \texttt{query\_card} and \texttt{query\_discard}, when card text, discard contents, attack effects, retreat costs, or other details are uncertain. Use game-action tools, such as \texttt{attack}, \texttt{play\_pokemon}, \texttt{evolve\_pokemon}, \texttt{attach\_energy}, \texttt{use\_supporter}, \texttt{use\_item}, \texttt{use\_tool}, \texttt{put\_stadium}, \texttt{retreat}, \texttt{use\_ability}, \texttt{choose\_card}, and \texttt{pass\_turn}, only when the corresponding action is legal in the current observation.
\end{quote}

\section{Implementation Details}
\label{sec:implementation_datails}
We used the OpenRouter API for model inference. Specifically, we set temperature to 0.7 and maximum output length to 2048 tokens. Unless otherwise specified, other API parameters were kept at their default values.

\end{document}